\documentclass{article}
\usepackage{ijcai21}

\usepackage{times}
\usepackage{soul}
\usepackage{url}
\usepackage[hidelinks]{hyperref}
\usepackage[utf8]{inputenc}
\usepackage[small]{caption}
\usepackage{graphicx}
\usepackage{amsmath}
\usepackage{amsthm}
\usepackage{booktabs}
\usepackage{algorithm}
\usepackage{algorithmic}
\usepackage{bbold}
\usepackage{listings}
\usepackage{xcolor}
\usepackage{tabularx}
\usepackage{booktabs}
\usepackage{multirow}
\usepackage{textcomp}
\usepackage{breqn}  
\usepackage{hyperref}
\usepackage{amssymb}
\usepackage{bbm}
\usepackage{bm}
\usepackage{subcaption}

\usepackage{caption}

\title{Time-Series Representation Learning via Temporal and Contextual Contrasting}

\author{
Emadeldeen Eldele$^1$ \and
Mohamed Ragab$^1$\and
Zhenghua Chen$^{2*}$\and
Min Wu$^{2}$\footnote{Corresponding Author}\and
Chee Keong Kwoh$^1$\and
Xiaoli Li$^2$\And
Cuntai Guan$^1$
\affiliations
$^1$School of Computer Science and Engineering, Nanyang Technological University, Singapore\\
$^2$Institute for Infocomm Research, A*STAR, Singapore\\
\emails
emad0002@ntu.edu.sg, \{mohamedr002,chen0832\}@e.ntu.edu.sg, wumin@i2r.a-star.edu.sg, asckkwoh@ntu.edu.sg, xlli@i2r.a-star.edu.sg, ctguan@ntu.edu.sg
}

\begin{document}
\thispagestyle{myheadings}
\markboth {}{This article has been published in the International Joint Conferences on Artificial Intelligence (IJCAI-21).}

\maketitle
\begin{abstract}
Learning decent representations from unlabeled time-series data with temporal dynamics is a very challenging task. In this paper, we propose an unsupervised \textbf{T}ime-\textbf{S}eries representation learning framework via \textbf{T}emporal and \textbf{C}ontextual \textbf{C}ontrasting (\textbf{TS-TCC}), to learn time-series representation from unlabeled data. First, the raw time-series data are transformed into two different yet correlated views by using weak and strong augmentations. Second, we propose a novel temporal contrasting module to learn \textit{robust} temporal representations by designing a tough cross-view prediction task. Last, to further learn \textit{discriminative} representations, we propose a contextual contrasting module built upon the contexts from the temporal contrasting module. It attempts to maximize the similarity among different contexts of the same sample while minimizing similarity among contexts of different samples. Experiments have been carried out on three real-world time-series datasets. The results manifest that training a linear classifier on top of the features learned by our proposed TS-TCC performs comparably with the supervised training. Additionally, our proposed TS-TCC shows high efficiency in few-labeled data and transfer learning scenarios. The code is publicly available at \url{https://github.com/emadeldeen24/TS-TCC}. 
\end{abstract}

\section{Introduction}
Time-series data are being incrementally collected on daily basis from IoT and wearable devices for various applications in healthcare, manufacturing, etc. However, they generally do not have human recognizable patterns and require specialists for annotation/labeling. Therefore, it is much harder to label time-series data than images, and little time-series data have been labeled in real-world applications \cite{ching2018opportunities}. Given that deep learning methods usually require a massive amount of labeled data for training, it is thus very challenging to apply them on time-series data with these labeling limitations. 

Self-supervised learning gained more attention recently to extract effective representations from unlabeled data for downstream tasks. Compared with models trained on full labeled data (i.e., supervised models), self-supervised pretrained models can achieve comparable performance with limited labeled data \cite{chen2020simple}. Various self-supervised approaches relied on different pretext tasks to train the models and learn representations from unlabeled data, such as solving puzzles \cite{puzzle} and predicting image rotation \cite{gidaris:unsupervised}. However, the pretext tasks can limit the generality of the learned representations. For example, classifying the different rotation angles of an image may deviate the model from learning features about the color or orientation of objects \cite{oord2018representation}.

Contrastive learning has recently shown its strong ability for self-supervised representation learning in computer vision domain because of its ability to learn invariant representation from augmented data  \cite{hjelm2018learning,He_2020_CVPR,chen2020simple}. It explores different views of the input images by first applying data augmentation techniques and then learns the representations by maximizing the similarity of different views from the same sample and minimizing the similarity with the views from different samples. However, these image-based contrastive learning methods are not able to work well on time-series data for the following reasons. First, they may not be able to address the temporal dependencies of data, which are key characteristics of time-series \cite{NEURIPS2019_53c6de78}. Second, some augmentation techniques used for images such as color distortion, generally cannot fit well with time-series data. So far, few works on contrastive learning have been proposed for time-series data. For example, \cite{mohsenvand20a,cheng2020subject} developed contrastive learning methods for bio-signals such as EEG and ECG. However, the above two methods are proposed for specific applications and they are not generalizable to other time-series data.

To address the above issues, we propose a \textbf{T}ime-\textbf{S}eries representation learning framework via \textbf{T}emporal and \textbf{C}ontextual \textbf{C}ontrasting (TS-TCC). Our framework employs simple yet efficient data augmentations that can fit any time-series data to create two different, but correlated views of the input data. Next, we propose a novel temporal contrasting module to learn robust representations by designing a tough cross-view prediction task, which for a certain timestep, it utilizes the past latent features of one augmentation to predict the future of another augmentation. This novel operation will force the model to learn robust representation by a harder prediction task against any perturbations introduced by different timesteps and augmentations. 
Furthermore, we propose a contextual contrasting module in TS-TCC to further learn discriminative representations upon the robust representations learned by the temporal contrasting module. 
In this contextual contrasting module, we aim to maximize the similarity among different contexts of the same sample while minimizing similarity among contexts of different samples.

In summary, the main contributions of this work are as follows.
\begin{itemize}
    \item A novel contrastive learning framework is proposed for unsupervised time-series representation learning. 
    
     \item Simple yet efficient augmentations are designed for time-series data in the contrastive learning framework. 
    
    \item We propose a novel temporal contrasting module to learn robust representations from time series data by designing a tough cross-view prediction task.  
    In addition, we propose a contextual contrasting module to further learn discriminative representations upon the robust representations.
    
    \item We perform extensive experiments on our proposed TS-TCC framework using three datasets. Experimental results show that the learned representations are effective for downstream tasks under supervised learning, semi-supervised learning and transfer learning settings.

\end{itemize}

\section{Related Works}

\subsection{Self-supervised Learning}
The recent advances in self-supervised learning started with applying pretext tasks on images to learn useful representations, such as solving jigsaw puzzles \cite{puzzle}, image colorization \cite{zhang2016colorful} and predicting image rotation \cite{gidaris:unsupervised}. Despite the good results achieved by these pretext tasks, they relied on heuristics that might limit the generality of the learned representations. On the other hand, contrastive methods started to shine via learning invariant representations from augmented data. For instance, MoCo \cite{He_2020_CVPR} utilized a momentum encoder to learn representations of negative pairs obtained from a memory bank. SimCLR \cite{chen2020simple} replaced the momentum encoder by using a larger batch of negative pairs. Also, BYOL \cite{grill2020bootstrap} learned representations by bootstrapping representations even without using negative samples. Last, SimSiam \cite{simsiam} supported the idea of neglecting the negative samples, and relied only on a Siamese network and stop-gradient operation to achieve the state-of-the-art performance.
While all these approaches have successfully improved representation
learning for visual data, they may not work well on time series data that have different properties, such as temporal dependency.

\subsection{Self-supervised Learning for Time-Series}
Representation learning for time series is becoming increasingly popular. Some approaches employed pretext tasks for time series data. For example, \cite{SSL_har} designed a binary classification pretext task for human activity recognition by applying several transformations on the data, and trained the model to classify between the original and the transformed versions. Similarly, SSL-ECG approach \cite{ecg_emotion_rec} learned ECG representations by applying six transformations to the dataset, and assigned pseudo labels according to the transformation type. 
Additionally, \cite{aggarwal2019adversarial} learned subject-invariant representations by modeling local and global activity patterns. Inspired by the success of contrastive learning, few works have recently leveraged contrastive learning for time series data. For example, CPC \cite{oord2018representation} learned representations by predicting the future in the latent space and showed great advances in various speech recognition tasks. Also, \cite{mohsenvand20a} designed EEG related augmentations and extended SimCLR model \cite{chen2020simple} to EEG data. Existing approaches used either temporal or global features. Differently, we first construct different views for input data by designing time-series specific augmentations. Additionally, we propose a novel cross-view temporal and contextual contrasting modules to improve the learned representations for time-series data.

\begin{figure}
\centering
\includegraphics[width=\columnwidth]{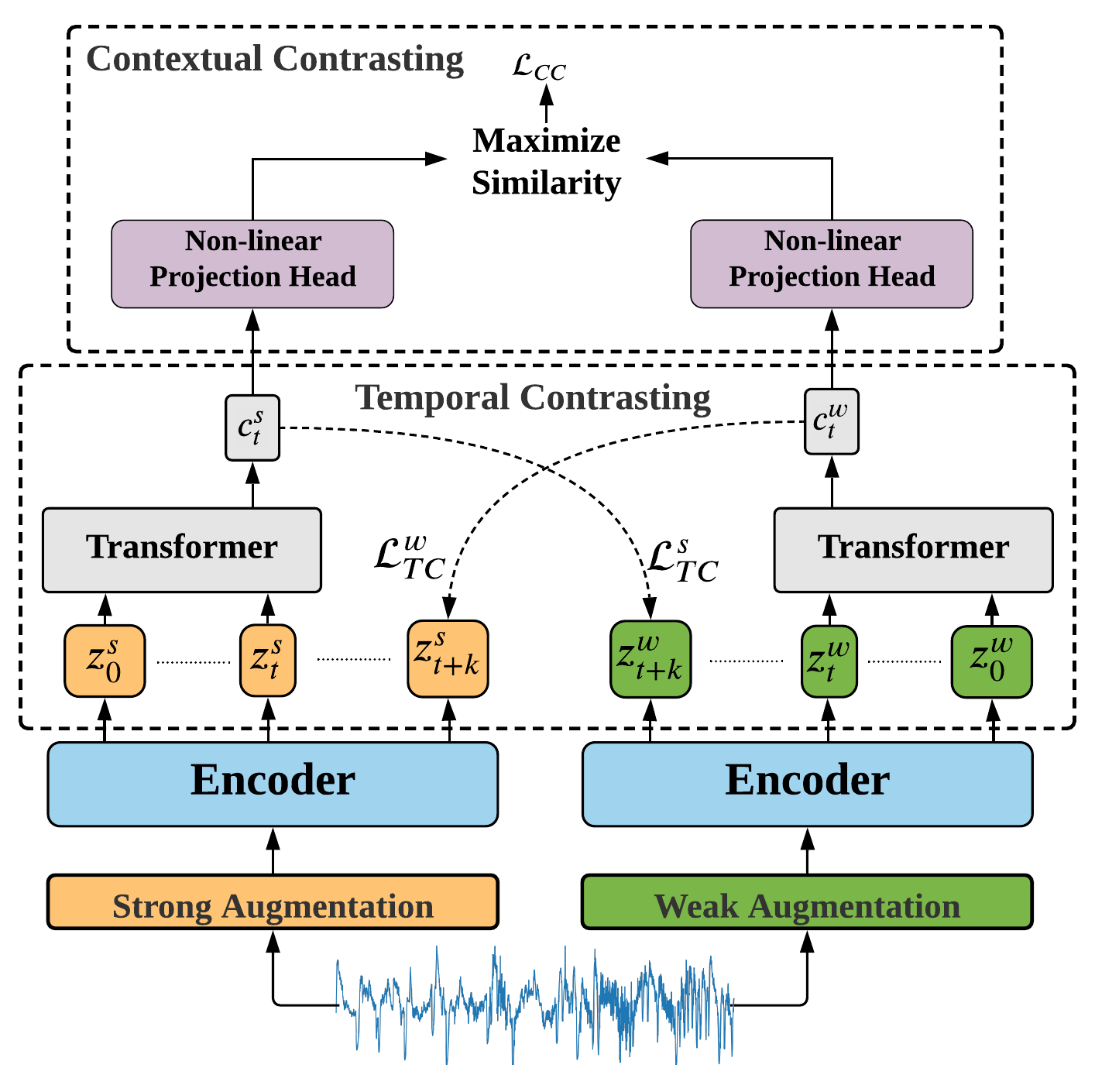}
\caption{Overall architecture of proposed TS-TCC model.}
\label{Fig:overall}
\end{figure}

\section{Methods}

This section describes our proposed TS-TCC in details. As shown in Figure~\ref{Fig:overall}, we first generate two different yet correlated views of the input data based on strong and weak augmentations. Then, a temporal contrasting module is proposed to explore the temporal features of the data with an autoregressive model. These models perform a tough cross-view prediction task by predicting the future of one view using the past of the other. We further maximize the agreement between the contexts of the autoregressive models by a contextual contrasting module. Next, we will introduce each component in the following subsections.

\subsection{Time-Series Data Augmentation}
Data augmentation is a key part in the success of the contrastive learning methods \cite{chen2020simple,grill2020bootstrap}. Contrastive methods try to maximize the similarity among different views of the same sample, while minimizing its similarity with other samples. It is thus important to design proper data augmentations for contrastive learning \cite{chen2020simple,mohsenvand20a}. Usually, contrastive learning methods use two (random) variants of the same augmentation. Given a sample $x$, they produce two views $x_1$ and $x_2$ sampled from the same augmentations family $\mathcal{T}$, i.e.,  $x_1 \!\sim\! \mathcal{T}$ and $x_2 \!\sim\! \mathcal{T}$. However, we argue that using different augmentations can improve the robustness of the learned representations. Consequently, we propose applying two separate augmentations, such that one augmentation is weak and the other is strong. In this paper, weak augmentation is a jitter-and-scale strategy. Specifically, we add random variations to the signal and scale up its magnitude. For strong augmentation, we apply permutation-and-jitter strategy, where permutation includes splitting the signal into a random number of segments with a maximum of $M$ and randomly shuffling them. Next, a random jittering is added to the permuted signal.
Notably, the augmentation hyperparameters should be chosen carefully according to the nature of the time-series data. For example, the value of $M$ in a time-series data with longer sequences should be greater than its value in those with shorter sequences when applying permutation. Similarly, the jittering ratio for normalized time-series data should be much less than the ratio for unnormalized data.

For each input sample $x$, we denote its strongly augmented view as $x^s$, and its weakly augmented view as $x^w$, where $x^s \!\sim\! \mathcal{T}_s$ and $x^w \!\sim\! \mathcal{T}_w$.
These views are then passed to the encoder to extract their high dimensional latent representations. In particular, the encoder has a 3-block convolutional architecture as proposed in \cite{wang2017time}. For an input $\mathbf{x}$, the encoder maps $\mathbf{x}$ into a high-dimensional latent representation $\mathbf{z} = f_{enc}(\mathbf{x})$.
We define $\mathbf{z} = [z_1, z_2, \dots z_{T}]$, where $T$ is the total timesteps, $z_i \in \mathbb{R}^{d}$, where $d$ is the feature length.
Thus, we get $\mathbf{z}^s$ for the strong augmented views, and $\mathbf{z}^w$ for the weak augmented views, which are then fed into the temporal contrasting module.

\subsection{Temporal Contrasting}
The Temporal Contrasting module deploys a contrastive loss to extract temporal features in the latent space with an autoregressive model.
Given the latent representations $\mathbf{z}$, the autoregressive model $f_{ar}$ summarizes all $\mathbf{z}_{\leq t}$ into a context vector $c_t = f_{ar}(\mathbf{z}_{\leq t}),~ c_t \in \mathbb{R}^{h}$, where $h$ is the hidden dimension of $f_{ar}$.
The context vector $c_t$ is then used to predict the timesteps from $z_{t+1}$ until $z_{t+k}$ $(1<k\leq K)$. To predict future timesteps, we use log-bilinear model that would preserve the mutual information between the input $x_{t+k}$ and $c_t$, such that $f_k(x_{t+k}, c_t) = exp((\mathcal{W}_k (c_t))^T z_{t+k})$, where $\mathcal{W}_k$ is a linear function that maps $c_t$ back into the same dimension as $z$, i.e. $\mathcal{W}_k : \mathbb{R}^{h \rightarrow d}$.

In our approach, the strong augmentation generates $c_t^s$ and the weak augmentation generates $c_t^w$. We propose a tough cross-view prediction task by using the context of the strong augmentation $c_t^s$ to predict the future timesteps of the weak augmentation $z_{t+k}^w$ and vice versa. The contrastive loss tries to minimize the dot product between the predicted representation and the true one of the same sample, while maximizing the dot product with the other samples $\mathcal{N}_{t, k}$ within the mini-batch.
Accordingly, we calculate the two losses $ \mathcal{L}_{TC}^s$ and $ \mathcal{L}_{TC}^w$ as follows:
\begin{equation}
    \mathcal{L}_{TC}^s=-\frac{1}{K} \sum_{k=1}^{K} \log \frac{\exp ((\mathcal{W}_k (c_t^s))^T z_{t+k}^{w})}{\sum_{n \in \mathcal{N}_{t, k}} \exp ((\mathcal{W}_k (c_t^s))^T z_{n}^w)}
\end{equation}
\begin{equation}
    \mathcal{L}_{TC}^w=-\frac{1}{K} \sum_{k=1}^{K} \log \frac{\exp ((\mathcal{W}_k (c_t^w))^T z_{t+k}^{s})}{\sum_{n \in \mathcal{N}_{t, k}} \exp ((\mathcal{W}_k (c_t^w))^T z_{n}^s)}
\end{equation}

\begin{figure}
\centering
\includegraphics[width=\columnwidth]{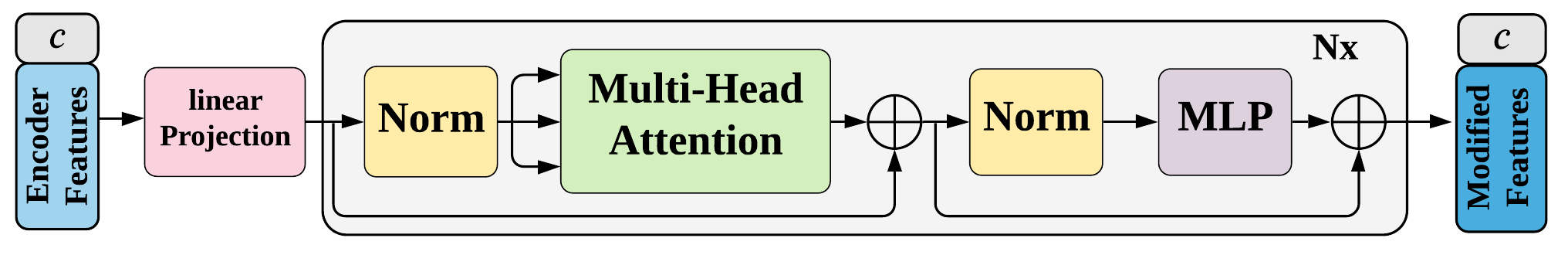}
\caption{Architecture of Transformer model used in Temporal Contrasting module. The token $c$ in the output is sent next to the Contextual Contrasting module.}
\label{Fig:transformer}
\end{figure}

We use Transformer as the autoregressive model because of its efficiency and speed \cite{NIPS2017_3f5ee243}. The architecture of the Transformer model is shown in Figure~\ref{Fig:transformer}. It mainly consists of successive blocks of multi-headed attention (MHA) followed by an MLP block. The MLP block is composed of two fully-connected layers with a non-linearity ReLU function and dropout in between. Pre-norm residual connections, which can produce more stable gradients \cite{wang_learning}, are adopted in our Transformer. We stack $L$ identical layers to generate the final features. Inspired by BERT model \cite{devlin2018bert}, we add a token $c \in \mathbb{R}^{h}$ to the input whose state acts as a representative context vector in the output. 
The operation of the Transformer starts by applying the features $\mathbf{z}_{\leq t}$ to a linear projection $\mathcal{W}_{Tran}$ layer that maps the features into the hidden dimension, i.e. $\mathcal{W}_{Tran} : \mathbb{R}^{d \rightarrow h}$. The output of this linear projection is then sent to the Transformer i.e. $\tilde{z} = \mathcal{W}_{Tran} (\mathbf{z}_{\leq t}),~~\tilde{\mathbf{z}} \in \mathbb{R}^{h}$.
Next, we attach the context vector into the features vector $\tilde{\mathbf{z}}$ such that the input features become $\psi_0 = [c; \tilde{\mathbf{z}}]$, where the subscript 0 denotes being the input to the first layer.
Next, we pass $\psi_0$ through Transformer layers as in the following equations:
\begin{align}
    \tilde{\psi_\ell} &= \operatorname{MHA}(\operatorname{Norm}(\psi_{\ell-1})) + \psi_{\ell-1}, && 1 \leq \ell \leq L; \label{eq:msa_apply} \\
    \psi_\ell &= \operatorname{MLP}(\operatorname{Norm}(\tilde{\psi_\ell})) + \tilde{\psi_\ell}, && 1 \leq \ell \leq L.  \label{eq:mlp_apply}
\end{align}
Finally, we re-attach the context vector from the final output such that $c_t = \psi_L^0$.
This context vector will be the input of the following contextual contrasting module.

\subsection{Contextual Contrasting}
We further propose a contextual contrasting module that aims to learn more discriminative representations. It starts with applying a non-linear transformation to the contexts using a non-linear projection head as in \cite{chen2020simple}. The projection head maps the contexts into the space of where the contextual contrasting is applied. 

Given a batch of $N$ input samples, we will have two contexts for each sample from its two augmented views, and thus have $2N$ contexts. 
For a context $c_t^i$, we denote $c_t^{i^+}$ as the positive sample of $c_t^{i}$ that comes from the other augmented view of the same input, and hence, ($c_t^i, c_t^{i^+}$) are considered to be a positive pair.  
Meanwhile, the remaining ($2N-2$) contexts from other inputs within the same batch are considered as the negative samples of $c_t^i$, i.e., $c_t^i$ can form ($2N-2$) negative pairs with its negative samples. Therefore, we can derive a contextual contrasting loss to maximize the similarity between the positive pair and minimizing the similarity between negative pairs. As such, the final representations can be discriminative. 

Eq. \ref{eqn:lcc} defines the contextual contrasting loss function $\mathcal{L}_{CC}$. Given a context $c_t^i$, we divide its similarity with its positive sample $c_t^{i^+}$ by its similarity with all the other ($2N - 1$) samples, including the positive pair and ($2N - 2$) negative pairs, to normalize the loss. 

\begin{equation}
    \mathcal{L}_{CC}= -\sum_{i=1}^{N} \log \frac{\exp \left(\operatorname{sim}\left(\boldsymbol{c}_t^{i}, \boldsymbol{c}_t^{i^+}\right) / \tau\right)}{\sum_{m=1}^{2 N} \mathbb{1}_{[m \neq i]} \exp \left(\operatorname{sim}\left(\boldsymbol{c}_t^{i}, \boldsymbol{c}_t^{m}\right) / \tau\right)},
    \label{eqn:lcc}
\end{equation}
where $sim(\bm u,\bm v) = \bm u^T \bm v/ \| \bm u\|\| \bm v\| $ denotes the dot product between $\ell_2$ normalized $\bm u$ and $\bm v$ (i.e., cosine similarity), $\mathbb{1}_{[m \neq i]} \in \{0,1\}$ is an indicator function, evaluating to $1$ iff $m \neq i$, and $\tau$ is a temperature parameter.

The overall self-supervised loss is the combination of the two temporal contrasting losses and the contextual contrasting loss as follows.
\begin{equation}
    \mathcal{L} = \lambda_1 \cdot (\mathcal{L}_{TC}^s + \mathcal{L}_{TC}^w ) + \lambda_2 \cdot \mathcal{L}_{CC},
    \label{eq:overall}
\end{equation}
where $\lambda_1$ and $\lambda_2$ are fixed scalar hyperparameters denoting the relative weight of each loss.

\section{Experimental Setup}
\subsection{Datasets}
To evaluate our model, we adopted three publicly available datasets for human activity recognition, sleep stage classification and epileptic seizure prediction, respectively. Additionally, we investigated the transferability of our learned features on a fault diagnosis dataset.

\subsubsection{Human Activity Recognition (HAR)}
We use UCI HAR dataset \cite{anguita2013public} which contains sensor readings for 30 subjects performing 6 activities (i.e. walking, walking upstairs, downstairs, standing, sitting, and lying down). They collected the data using a mounted Samsung Galaxy S2 device on their waist, with a sampling rate of 50 Hz.

\subsubsection{Sleep Stage Classification}
In this problem, we aim to classify the input EEG signal into one of five classes:  Wake (W), Non-rapid eye movement (N1, N2, N3) and Rapid Eye Movement (REM).
We downloaded Sleep-EDF dataset from the PhysioBank \cite{goldberger2000physiobank}. Sleep-EDF includes whole-night PSG sleep recordings, where we used a single EEG channel (i.e., Fpz-Cz) with a sampling rate of 100 Hz, following previous studies \cite{attnSleep_paper}.

\subsubsection{Epilepsy Seizure Prediction}
The Epileptic Seizure Recognition dataset \cite{PhysRevE.64.061907} consists of EEG recordings from 500 subjects, where the brain activity was recorded for each subject for 23.6 seconds. Note that the original dataset is labeled with five classes. As four of them do not include epileptic seizure, so we merged them into one class and treat it as a binary classification problem.

\begin{table}[]
\centering
\resizebox{0.47\textwidth}{!}{
\begin{tabular}{@{}l|ccccc@{}}
\toprule
Dataset & \# Train & \# Test & Length & \# Channel & \# Class \\ \midrule
HAR & 7352 & 2947 & 128 & 9 & 6 \\
Sleep-EDF & 25612 & 8910 & 3000 & 1 & 5 \\
Epilepsy & 9200 & 2300 & 178 & 1 & 2 \\ 
FD & 8184 & 2728 & 5120 & 1 & 3 \\ \bottomrule
\end{tabular}
}
\caption{Description of datasets used in our experiments. The details of FD is the same for all the 4 working conditions.}
\label{tab:data}
\end{table}

\subsubsection{Fault Diagnosis (FD)}
We conducted the transferability experiment on a real-world fault diagnosis dataset \cite{lessmeier2016condition}. This dataset was collected under four different working conditions. Each working condition can be considered as a separate domain as it has different characteristics from the other working conditions \cite{ragab2020adversarial}.
Each domain has three classes, namely, two fault classes (i.e., inner fault and outer fault) and one healthy class.

Table \ref{tab:data} summarizes the details of each dataset, e.g., the number of training samples (\# Train) and testing samples (\# Test), the length of the sample, the number of sensor channels (\# Channel) and the number of classes (\# Class).

\begin{table*}
\centering

\begin{tabular}{@{}l|cc|cc|cc@{}}
\toprule
 & \multicolumn{2}{c|}{HAR} & \multicolumn{2}{c|}{Sleep-EDF} & \multicolumn{2}{c}{Epilepsy} \\ \midrule
Baseline       & ACC           & MF1           & ACC           & MF1           & ACC           & MF1           \\ \midrule

Random Initialization  & 57.89$\pm$5.13 & 55.45$\pm$5.49 & 35.61$\pm$6.96 & 23.80$\pm$7.96 & 90.26$\pm$1.77 & 81.12$\pm$4.22 \\ 

Supervised      & 90.14$\pm$2.49 & 90.31$\pm$2.24 & \textbf{83.41$\pm$1.44} & \textbf{74.78$\pm$0.86} & 96.66$\pm$0.24 & 94.52$\pm$0.43 \\

SSL-ECG \cite{ecg_emotion_rec}  & 65.34$\pm$1.63 & 63.75$\pm$1.37 & 74.58$\pm$0.60 & 65.44$\pm$0.97 & 93.72$\pm$0.45 & 89.15$\pm$0.93 \\

CPC \cite{oord2018representation} & 83.85$\pm$1.51 & 83.27$\pm$1.66 & 82.82$\pm$1.68 & 73.94$\pm$1.75 & 96.61$\pm$0.43 & 94.44$\pm$0.69 \\ 

SimCLR \cite{chen2020simple} & 80.97$\pm$2.46 & 80.19$\pm$2.64 & 78.91$\pm$3.11 & 68.60$\pm$2.71  & 96.05$\pm$0.34 & 93.53$\pm$0.63 \\

TS-TCC \textit{(ours)} & \textbf{90.37$\pm$0.34} & \textbf{90.38$\pm$0.39} & 83.00$\pm$0.71 & 73.57$\pm$0.74 & \textbf{97.23$\pm$0.10} & \textbf{95.54$\pm$0.08} \\ 
\bottomrule

\end{tabular}
\caption{Comparison between our proposed TS-TCC model against baselines using linear classifier evaluation experiment.}

\label{tbl:fine_linear}
\end{table*}

\subsection{Implementation Details}
We split the data into 60\%, 20\%, 20\% for training, validation and testing, with considering subject-wise split for Sleep-EDF dataset to avoid overfitting.
Experiments were repeated for 5 times with 5 different seeds, and we reported the mean and standard deviation.
The pretraining and downstream tasks were done for 40 epochs, as we noticed that the performance does not improve with further training.
We applied a batch size of 128 (which was reduced to 32 in \textit{few-labeled data} experiments as data size may be less than 128).
We used Adam optimizer with a learning rate of 3e-4, weight decay of 3e-4, $\beta_1 = 0.9$, and $\beta_2 = 0.99$.
For the strong augmentation, we set $M_{HAR}=10,~M_{Ep}=12$ and $M_{EDF}=20$, while for the weak augmentation, we set the scaling ratio to 2 for all the datasets.
We set $\lambda_1=1$, while we achieved good performance when $\lambda_2 \approx 1$. Particularly, we set it as 0.7 in our experiments on the four datasets.
In the Transformer, we set the $L=4$, and the number of heads as 4. We tuned $h \in \{32, 50, 64, 100, 128, 200, 256\}$ and set $h_{HAR,Ep}=100$, $h_{EDF}=64$. We also set its dropout to 0.1.
In contextual contrasting, we set $\tau=0.2$.
Lastly, we built our model using PyTorch 1.7 and trained it on a NVIDIA GeForce RTX 2080 Ti GPU.

\section{Results}
To show the efficacy of our proposed TS-TCC, we test it on three different training settings, including linear evaluation, semi-supervised training and transfer learning. 
We evaluate the performance using two metrics namely the accuracy and the macro-averaged F1-score (MF1) to better evaluate the imbalanced datasets.

\subsection{Comparison with Baseline Approaches}
We compare our proposed approach against the following baselines. 
(1) \textbf{Random Initialization}: training a linear classifier on top of randomly initialized encoder; (2) \textbf{Supervised}: supervised training of both encoder and classifier model; (3) \textbf{SSL-ECG} \cite{ecg_emotion_rec};
(4) \textbf{CPC} \cite{oord2018representation}; (5)  \textbf{SimCLR} \cite{chen2020simple}. It is worth noting that, we use time-series specific augmentations to adapt SimCLR to our application as it was originally designed for images.

To evaluate the performance of our TS-TCC model, we follow the standard linear benchmarking evaluation scheme~\cite{oord2018representation,chen2020simple}. Particularly,  we train a linear classifier (single MLP layer) on top of a frozen self-supervised pretrained encoder model. Table \ref{tbl:fine_linear} shows the linear evaluation results of our approach against the baseline methods. Overall, our proposed TS-TCC outperforms all the three state-of-the-art methods. Furthermore, TS-TCC, with only linear classifier, performs best on two out of three datasets while achieving comparable performance to the supervised approach on the third dataset. This demonstrates the powerful representation learning capability of our TS-TCC model. Notably, contrastive methods (e.g., CPC, SimCLR and our TS-TCC) generally achieve better results than the pretext-based method (i.e., SSL-ECG), which reflects the power of invariant features learned by contrastive methods. Additionally, CPC method shows better results than SimCLR, indicating that temporal features are more important than general features in time-series data.

\subsection{Semi-supervised Training}
\begin{figure}
\centering
\includegraphics[width=\columnwidth]{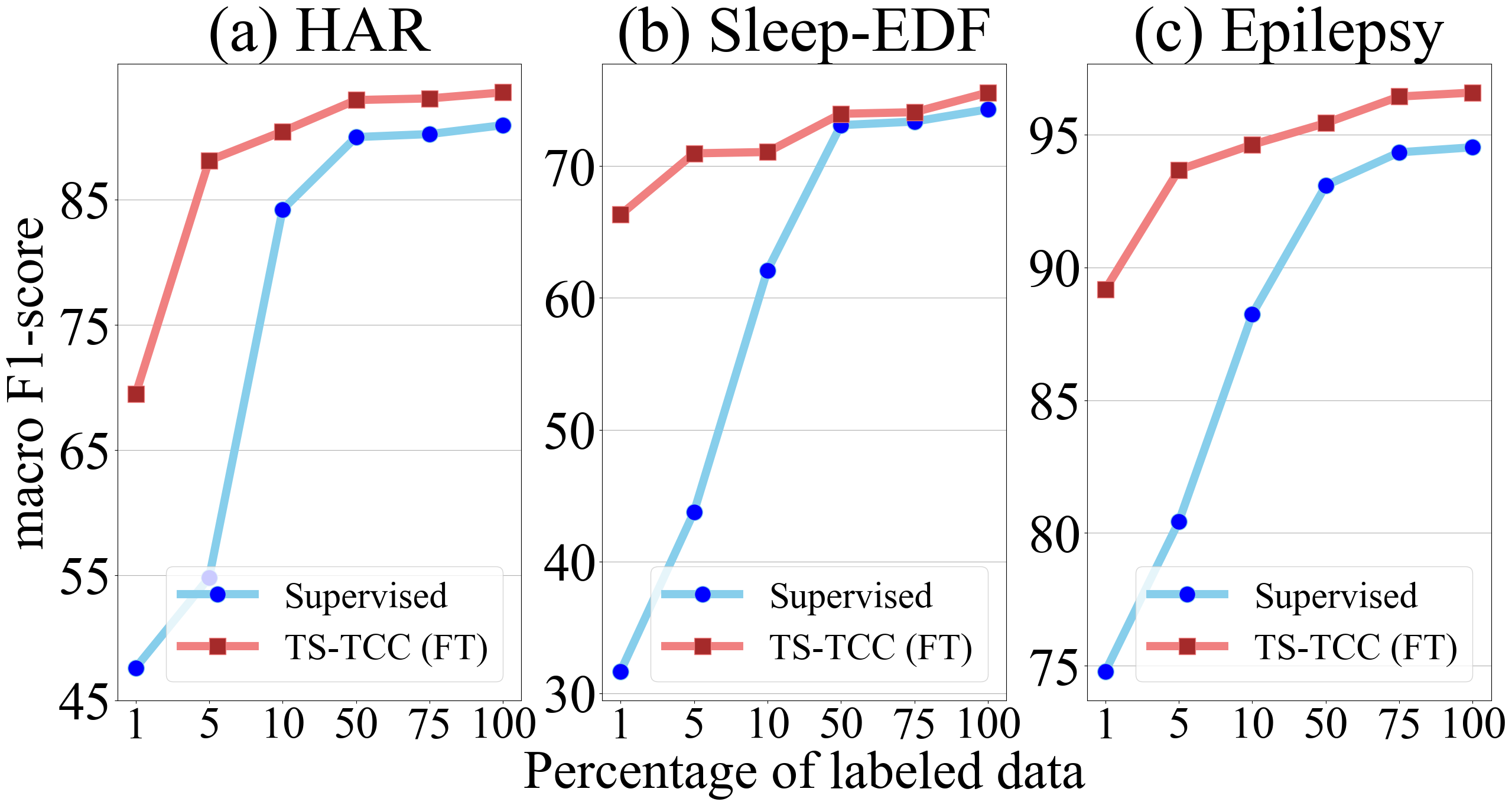}
\caption{Comparison between supervised training vs. TS-TCC fine-tuning for different few-labeled data scenarios in terms of MF1.}
\label{Fig:few_labels}
\end{figure}

We investigate the effectiveness of our TS-TCC under the semi-supervised settings, by training the model with  1\%, 5\%, 10\%, 50\%, and 75\% of randomly selected instances of the training data.
Figure~\ref{Fig:few_labels} shows the results of our TS-TCC along with the supervised training under the aforementioned settings. In particular, TS-TCC fine-tuning (i.e., red curves in Figure~\ref{Fig:few_labels}) means that we fine-tuned the pretrained encoder with few labeled samples. 

We observe that supervised training performs poorly with limited labeled data, while our TS-TCC fine-tuning achieves significantly better performance than supervised training with only 1\% of labeled data. For example, TS-TCC fine-tuning can still achieve around 70\% and 90\% for HAR and Epilepsy datasets respectively. Furthermore, our TS-TCC fine-tuning with only 10\% of labeled data can achieve comparable performance with the supervised training with 100\% of labeled data in the three datasets, demonstrating the effectiveness of our TS-TCC method under the semi-supervised setting. 

\begin{table*}[!h]
\centering
 \resizebox{!}{0.77cm}{
\begin{tabular}{@{}c|cccccccccccc|c@{}}
\toprule
\textbf{} & A$\rightarrow$B & A$\rightarrow$C & A$\rightarrow$D & B$\rightarrow$A & B$\rightarrow$C & B$\rightarrow$D & C$\rightarrow$A & C$\rightarrow$B & C$\rightarrow$D & D$\rightarrow$A & D$\rightarrow$B & D$\rightarrow$C & AVG \\ \midrule
Supervised & 34.38 & 44.94 & 34.57 & \textbf{52.93} & 63.67 & \textbf{99.82} & \textbf{52.93} & 84.02 & 83.54 & \textbf{53.15} & 99.56 & 62.43 & 63.83 \\
TS-TCC \textit{(FT)} & \textbf{43.15} & \textbf{51.50} & \textbf{42.74} & 47.98 & \textbf{70.38} & 99.30 & 38.89 & \textbf{98.31} & \textbf{99.38} & 51.91 & \textbf{99.96} & \textbf{70.31} & \textbf{67.82} \\ \hline
\end{tabular}
}
\caption{Cross-domains transfer learning experiment applied on Fault Diagnosis dataset in terms of accuracy. (FT stands for fine-tuning)}
\label{tbl:TL}
\end{table*}

\begin{table*}[!htb]
\centering

\begin{tabular}{@{}l|cc|cc|cc@{}}
\toprule
 & \multicolumn{2}{c|}{HAR} & \multicolumn{2}{c|}{Sleep-EDF} & \multicolumn{2}{c}{Epilepsy} \\ \midrule
Component & ACC & MF1 & ACC & MF1 & ACC & MF1 \\ \midrule

TC only  & 82.76$\pm$1.50 & 82.17$\pm$1.64 & 80.55$\pm$0.39 & 70.99$\pm$0.86 & 94.39$\pm$1.19 & 90.93$\pm$1.41 \\

TC + X-Aug & 87.86$\pm$1.33 & 87.91$\pm$1.09 & 81.58$\pm$1.70 & 71.88$\pm$1.71 & 95.56$\pm$0.24 & 92.57$\pm$0.29 \\ 

TS-TCC (TC + X-Aug + CC)  & \textbf{90.37$\pm$0.34} & \textbf{90.38$\pm$0.39} & \textbf{83.00$\pm$0.71} & \textbf{73.57$\pm$0.74} & \textbf{97.23$\pm$0.10} & \textbf{95.54$\pm$0.08} \\

\midrule

TS-TCC (Weak only)   & 76.55$\pm$3.59 & 75.14$\pm$4.66 & 80.90$\pm$1.87 & 72.51$\pm$1.74 & 97.18$\pm$0.17 & 95.47$\pm$0.31 \\ 

TS-TCC (Strong only)  & 60.23$\pm$3.31 & 56.15$\pm$4.14 & 78.55$\pm$2.94 & 68.05$\pm$1.87 & 97.14$\pm$0.23 & 95.39$\pm$0.29 \\

\bottomrule
\end{tabular}
\caption{Ablation study of each component in TS-TCC model performed with linear classifier evaluation experiment.}
\label{tbl:ablation}
\end{table*}

\subsection{Transfer Learning Experiment}
We further examine the transferability of the learned features by designing a transfer learning experiment. We use Fault Diagnosis (FD) dataset introduced in Table \ref{tab:data} for the evaluation under the transfer learning setting. Here, we train the model on one condition (i.e., source domain) and test it on another condition (i.e., target domain). In particular, we adopt two training schemes on the source domain, namely, (1) supervised training and (2) TS-TCC fine-tuning where we fine-tuned our pretrained encoder using the labeled data in the source domain.  

Table \ref{tbl:TL} shows the performance of the two training schemes under 12 cross-domain scenarios. Clearly, our pretrained TS-TCC model with fine-tuning (FT) consistently outperforms the supervised pretraining in 8 out of 12 cross-domain scenarios. TS-TCC model can achieve at least 7\% improvement in 7 out of 8 winning scenarios (except for D$\rightarrow$B scenario). Overall, our proposed approach can improve the transferability of learned representations over the supervised training by about 4\% in terms of accuracy.

\subsection{Ablation Study}
We study the effectiveness of each component in our proposed TS-TCC model.
Specifically, we derive different model variants for comparison as follows. First, we train the Temporal Contrasting module (TC) without the cross-view prediction task, where each branch predicts the future time-steps of the same augmented view. This variant is denoted as `TC only'. Second, we train the TC module with adding the cross-view prediction task, which is denoted as `TC + X-Aug'. Third, we train the whole proposed TS-TCC model, which is denoted as `TC + X-Aug + CC'. We also study the effect of using a single augmentation in TS-TCC. In particular, for an input $x$, we generate two different views $x_1$ and $x_2$ from the same augmentation type, i.e., $x_1 \!\sim\! \mathcal{T}_w$ and $x_2 \!\sim\! \mathcal{T}_w$ when using the weak augmentation.

Table \ref{tbl:ablation} shows this ablation study on the three datasets. 
Clearly, the proposed cross-view prediction task generates robust features and thus improves the performance by more than 5\% on HAR datasets, and $\sim$1\% on Sleep-EDF and Epilepsy datasets. Additionally, the contextual contrasting module further improves the performance, as it helps the features to be more discriminative. Studying the augmentations effect, we find that generating different views from the same augmentation type is not helpful with HAR and Sleep-EDF datasets. On the other hand, Epilepsy dataset can achieve comparable performance with only one augmentation. Overall, our proposed TS-TCC method using both types of augmentations achieves the best performance.

\subsection{Sensitivity Analysis}
\label{sec:sens_analysis}
We perform sensitivity analysis on HAR dataset to study three parameters namely, the number of predicted future timesteps $K$ in the temporal contrasting module, besides $\lambda_1$ and $\lambda_2$ in Eq. \ref{eqn:lcc}.

\begin{figure*}
     \centering
     \begin{subfigure}[b]{0.3\textwidth}
         \centering
         \includegraphics[width=\textwidth]{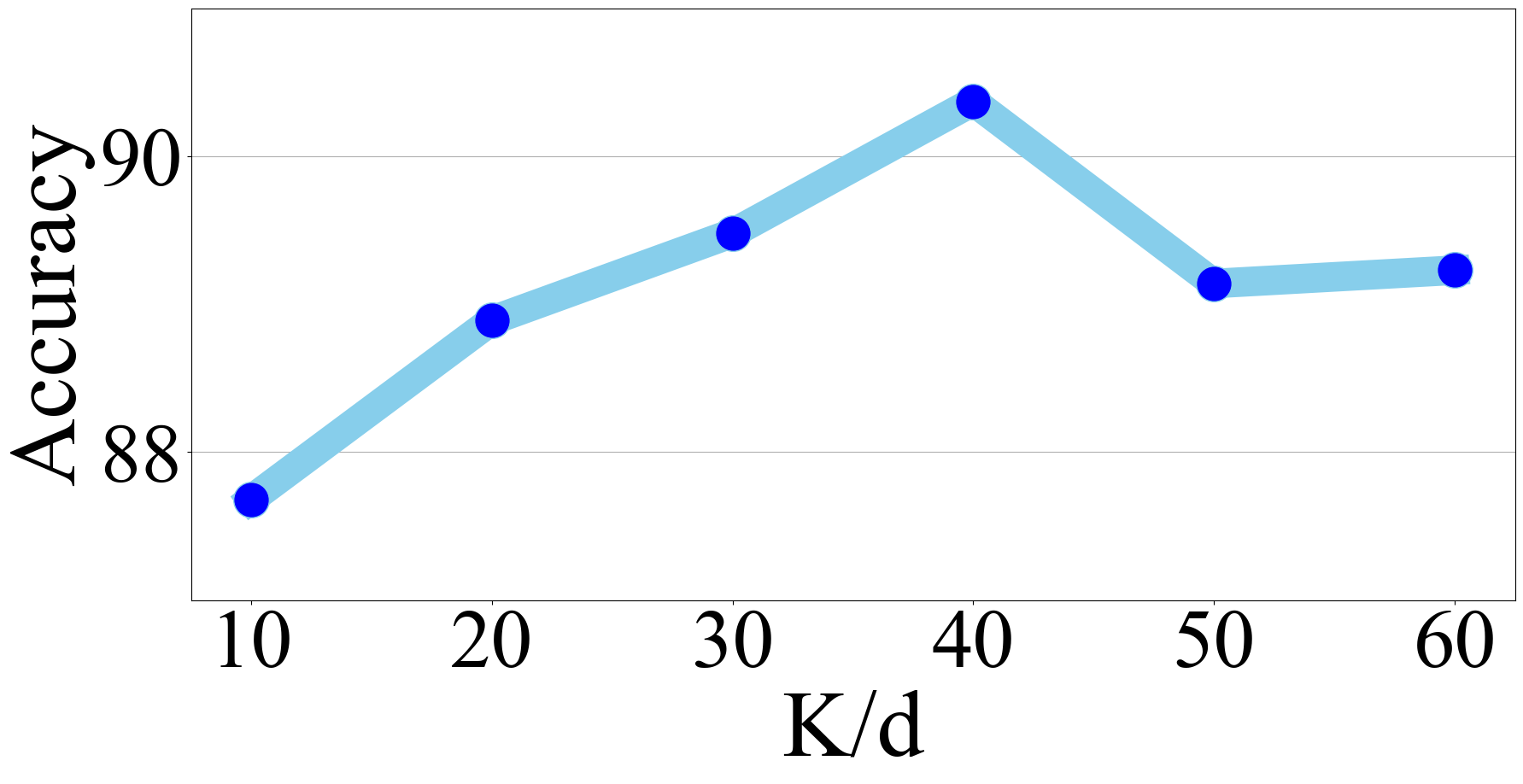}
         \caption{}
         \label{fig:sens1}
     \end{subfigure}
     \hfill
     \begin{subfigure}[b]{0.3\textwidth}
         \centering
         \includegraphics[width=\textwidth]{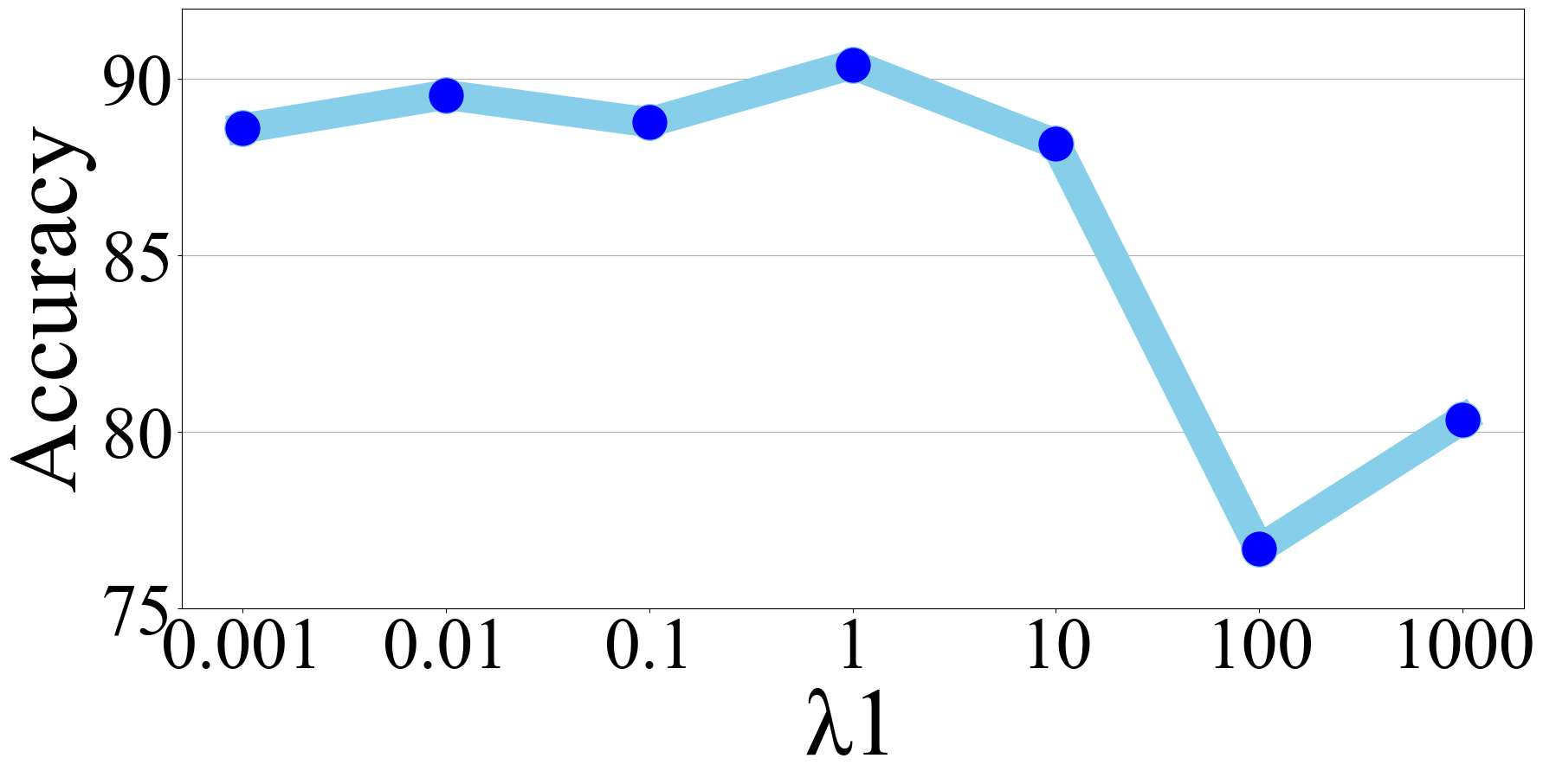}
         \caption{}
         \label{fig:sens2}
     \end{subfigure}
     \hfill
     \begin{subfigure}[b]{0.3\textwidth}
         \centering
         \includegraphics[width=\textwidth]{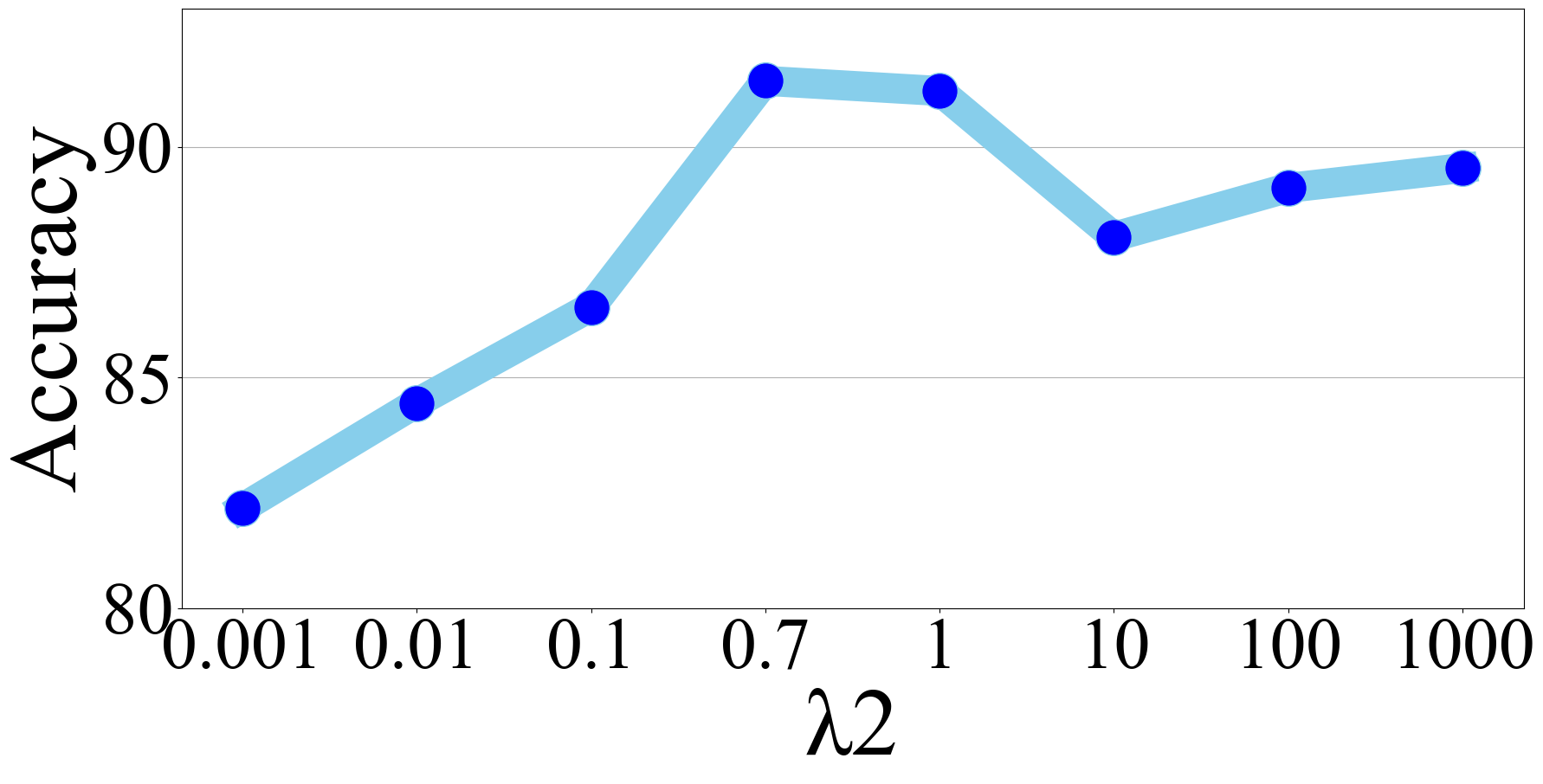}
         \caption{}
         \label{fig:sens3}
     \end{subfigure}
        \caption{Three sensitivity analysis experiments on HAR dataset.}
        \label{Fig:ts_sens_analysis}
\end{figure*}

Figure~\ref{fig:sens1} shows the effect of $K$ on the overall performance, where x-axis is the percentage $K/d$, $d$ is the length of the features. Clearly, increasing the percentage of the predicted future timesteps improves the performance. However, larger percentages can harm the performance as it reduces the amount of past data used for training the autoregressive model. We observe that predicting 40\% of the total feature length performs the best, and thus we set $K$ as $d\times$40\% in our experiments. Figures~\ref{fig:sens2} and \ref{fig:sens3} show the results of varying $\lambda_1$ and $\lambda_2$ in a range between 0.001 and 1000 respectively. We fix $\lambda_1=1$ and change the values of $\lambda_2$ in Figure~\ref{fig:sens3}. We observe that our model achieves good performance when $\lambda_2 \approx 1$, where the model performs best with $\lambda_2=0.7$. 
Consequently, we fix $\lambda_2=0.7$ and tune the value of $\lambda_1$ as in Figure~\ref{fig:sens2}, where we find that our model achieves the best performance when $\lambda_1=1$. We also find that as $\lambda_1<10$, our model is less sensitive to its value, while it is more sensitive to different values of $\lambda_2$.



\section{Conclusions}
We propose a novel framework called TS-TCC for unsupervised representation learning from time-series data. 
The proposed TS-TCC framework first creates two views for each sample by applying strong and weak augmentations. Then the temporal contrasting module learns robust temporal features by applying a tough cross-view prediction task. We further propose a contextual contrasting module to learn discriminative features upon the learned robust representations. The experiments show that a linear classifier trained on top the features learned by our TS-TCC performs comparably with supervised training. In addition, our proposed TS-TCC shows high efficiency on few-labeled data and transfer learning scenarios, e.g., our TS-TCC by using only 10\% of the labeled data can achieve close performance to the supervised training with full labeled data.

\section*{Acknowledgements}
This research is supported by the Agency for Science, Technology and Research (A*STAR) under its AME Programmatic Funds (Grant No. A20H6b0151) and Career Development Award (Grant No. C210112046).

\bibliographystyle{named}
\bibliography{ijcai21}

\end{document}